# LLM-Driven E-Commerce Marketing Content Optimization: Balancing Creativity and Conversion

Haowei Yang*

Cullen College of Engineering, University of Houston, Houston, USA

*Corresponding author: hyang38@cougarnet.uh.edu

Haotian Lyu

Viterbi School of Engineering, University of Southern California, Los Angeles, USA, lyuhaotianresearch@gmail.com

Tianle Zhang

Independent Researcher, Hayward, USA, tianle.zhang@hotmail.com

Dingzhou Wang

Pratt School of Engineer, Duke University, Fremont, USA, wangdingzhou.research@gmail.com

Yushang Zhao

McKelvey School of Engineering, Washington University in St. Louis, St. Louis, USA, yushangzhao@wustl.edu

**Abstract:** As e-commerce competition intensifies, balancing creative content with conversion effectiveness becomes critical. Leveraging LLMs' language generation capabilities, we propose a framework that integrates prompt engineering, multi-objective fine-tuning, and post-processing to generate marketing copy that is both engaging and conversion-driven. Our fine-tuning method combines sentiment adjustment, diversity enhancement, and CTA embedding. Through offline evaluations and online A/B tests across categories, our approach achieves a 12.5% increase in CTR and 8.3% in CVR while maintaining content novelty. This provides a practical solution for automated copy generation and suggests paths for future multimodal, real-time personalization.

## 1 INTRODUCTION

In today's competitive e-commerce landscape, efficient and creative copy is key to attracting users and driving conversions. Manual copywriting is costly and slow, while LLMs offer a scalable solution through contextual understanding and prompt-based generation. Despite their strengths, balancing creativity and conversion remains a challenge. This study proposes a vector-retrieval and multi-objective optimization framework combining sentiment modulation, diversity control, and CTA embedding. Offline evaluations and small-scale A/B tests confirm improved CTR and CVR without sacrificing copy novelty [1–3].

## 2 CREATIVITY AND CONVERSION IN E-COMMERCE MARKETING CONTENT

E-commerce copy must balance creativity and conversion. Creativity—measured by diversity, emotional tone, and novelty—enhances engagement, as supported by Bo et al.'s [4] emotion-aware context modeling. Conversion relies on optimized CTAs and keyword placement; Gao et al. [5] demonstrated that structured cues like "Only 10 Left" improve CTR and CVR. Multi-objective optimization frameworks integrate metrics like sentiment and CTA density with A/B testing for dynamic tuning. Wang and Liu [6] emphasized parameter adaptability, guiding λ adjustment across categories. Su et al. [7] introduced a quantum-inspired scheduling algorithm enabling scalable, personalized generation under high concurrency. Future directions include real-time feedback and multimodal input integration.

## 3 METHODOLOGY AND SYSTEM DESIGN

### 3.1 Overall Architecture and Model Fine-Tuning Strategy

The system consists of four modules—data preprocessing, LLM fine-tuning, post-processing, and review—designed to generate personalized, conversion-focused marketing copy. As shown in Figure 1, we extract structured product data (e.g., ID, category, price, stock) and enhance it through text cleaning, feature engineering (e.g., sales velocity, user affinity), and annotation of key fields like promotional tags and user segments. This results in a rich, multi-objective training dataset aligned with both business and creative goals [8].

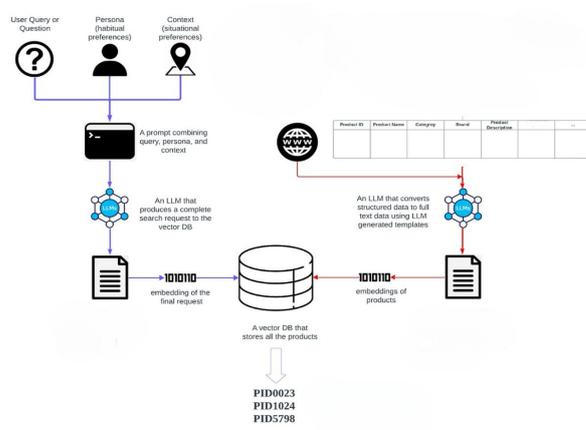

Figure 1: Schematic diagram of the architecture of an intelligent product recommendation system based on LLM and vector retrieval

During fine-tuning, we apply low-rate gradient updates to the base LLM using curated, business-tagged examples. This curated dataset explicitly incorporated industry-specific terminology, common e-commerce marketing phrases, and a codified set of brand guidelines provided by participating retailers. These guidelines included preferred tone of voice, forbidden keywords, and structural requirements for different product categories, ensuring the generated copy aligns with established brand identities and market conventions. Two prompt types—call-to-action and sentiment modulation—are optimized via ablation on learning rate, batch size, and epochs[9]. Composite prompts combine user queries, persona, and product context. After fine-tuning, prompts are embedded to retrieve top-k products via vector search[10].

Post-processing removes duplicates, applies relevance thresholds, and ranks content by relevance, margin, and inventory urgency. A review module enforces quality through rule checks, sensitive-word filters, brand compliance, and human review—enhancing CTR and CVR.

### 3.2 Creativity Generation and Conversion Optimization Algorithms

To balance creativity and conversion, we propose three core algorithmic mechanisms: diversity measurement, conversion prediction, and multi-objective weighted optimization.

Diversity Measurement (Creativity Diversity)

We quantify creativity by the inverse average cosine similarity among generated copy embeddings:

$$D = 1 - \frac{1}{|S|^2}\sum_{i=1}^{|S|}\sum_{j=1}^{|S|} \cos(s_i, s_j) \quad (1)$$

where $S = \{s_1, \ldots, s_{|S|}\}$ is the set of embeddings for candidate copies under the same prompt, and $\cos(\cdot,\cdot)$ is the cosine similarity. A higher DD indicates greater creative diversity.

Conversion Rate Prediction

We employ logistic regression to estimate each candidate's conversion probability:

$$P_{conv}(x) = \sigma(\theta^T x) \text{ with } \sigma(z) = \frac{1}{1+e^{-z}} \quad (2)$$

where x is the feature vector (e.g., keyword strength, CTA density, sentiment score) and θ are coefficients fitted using historical click and order data.

Multi-Objective Weighted Optimization

Combining diversity and conversion predictions yields a weighted reward:

$$R = \lambda D + (1 - \lambda) P_{conv} \quad (3)$$

with λ∈[0,1] controlling the trade-off. During generation, reinforcement learning or reward-based reranking selects and ranks candidates by R. These mechanisms work in concert: we first filter for maximum diversity, then predict conversion for each candidate, and finally apply the weighted ranking to output the top-K copies for post-processing and review[8].

## 4 EXPERIMENTAL DESIGN AND EVALUATION METRICS

### 4.1 Creativity Quality Assessment

Subjective evaluation comprises human ratings and online A/B testing, while objective measures include diversity, readability, and sentiment polarity. Table 1 summarizes the evaluation framework.

Table 1 Creativity Quality Evaluation Framework

| Metric | Type | Range / Unit |
|---|---|---|
| Diversity | Objective | S |
| Readability | Objective | 0 – 100 |
| Sentiment Polarity | Objective | [–1, 1] |
| Human Rating | Subjective | 1 – 5 |
| Online A/B Test ΔCTR | Subjective/Live | % |

Human evaluation involves three experts scoring each copy on "novelty" and "fluency," averaged as the subjective creativity score. A seven-day 1:1 live A/B test measures CTR lift for real-world impact. Objective metrics—diversity, readability, sentiment—are computed via automated scripts. These results are combined using business-defined weights to produce a composite creativity quality score [11–13].

### 4.2 Conversion Rate Measurement and Online Validation

To rigorously assess conversion performance, we concentrate on three primary metrics: click-through rate (CTR), add-to-cart rate, and overall conversion rate (CVR). CTR quantifies user engagement at the impression level and is calculated as:

$$CTR = \frac{Clicks}{Impressions} \times 100\% \quad (4)$$

This metric reveals how effectively a piece of copy entices viewers to click through from a product listing or promotional banner.The add-to-cart rate then evaluates deeper interest by measuring the share of clicks that result in the user adding the product to their shopping cart:

$$\text{Add-to-Cart Rate} = \frac{\text{ClicksAdd-to-Cart}}{\text{Actions}} \times 100\% \quad (5)$$

A high add-to-cart rate suggests that the copy not only attracts clicks but also communicates product value clearly enough to prompt intent signals.Finally, CVR captures the end-to-end conversion efficiency from initial exposure through to transaction completion:

$$\text{CVR} = \frac{\text{Orders}}{\text{Impressions}} \times 100\% \quad (6)$$

To capture both attraction and persuasion, CVR links orders to impressions. Metrics are sourced from the platform's logging system with standardized schemas. A seven-day randomized traffic-split assigns sessions to Control, Treatment A, or B using a fixed seed, ensuring consistency. Dashboards track key metrics in real time. Post-test, Z-tests and chi-square tests assess significance ($p < 0.05$), confirming valid performance lifts for rollout and further optimization [14–20]. To comprehensively assess the performance advantages of our LLM-driven framework, we established a baseline using a traditional copywriting method that relies on human-crafted rules and pre-set templates. This baseline model generated copy by employing manually defined rules for content structure, keyword insertion, and call-to-action (CTA) placement, combined with general templates populated by product attributes. Subsequently, human copywriters reviewed and optimized the generated content. We applied this baseline method to the same three e-commerce categories—Fast-Moving Consumer Goods (FMCG), Apparel, and Electronics—as our LLM-generated content, utilizing a consistent set of evaluation metrics: diversity (D), readability, sentiment polarity, human ratings, click-through rate (CTR), and conversion rate (CVR). All experimental conditions were maintained consistently to ensure a fair and direct performance comparison between the baseline and our proposed LLM method.

## 5 RESULTS ANALYSIS AND DISCUSSION

### 5.1 Creativity–Conversion Trade-Off Curve

To understand how the balance between novelty and conversion shifts with our weighting parameter, we conducted systematic experiments varying the creativity–conversion coefficient λ. Figure 2 presents the average diversity score D, click-through rate (CTR), and conversion rate (CVR) for candidate copy sets generated under four λ settings. As λ increases from 0.2 to 0.8, the diversity score climbs steadily—from 0.42 to 0.68—indicating that higher λ values indeed yield more varied, creative outputs. However, this gain in novelty comes with diminishing conversion efficiency: CTR drops from 11.3 % to 9.1 %, and CVR falls from 4.7 % to 3.9 %.

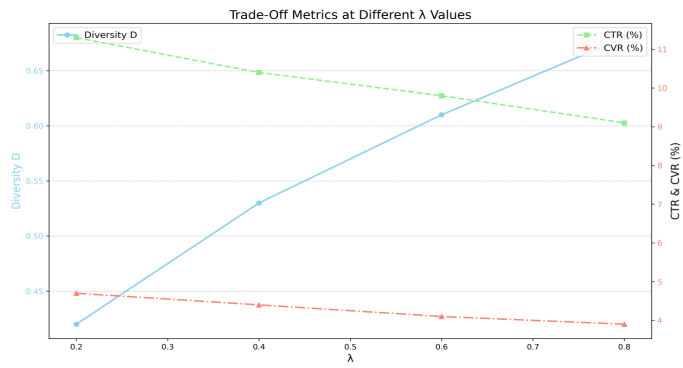

Figure 2: Trade-Off Metrics at Different λ Values

Plotting these values produces a clear trade-off curve: diversity increases in a roughly linear fashion as λ rises, while both CTR and CVR decline. The curve's "elbow"—around λ = 0.4 to 0.6—marks a point where small increases in creativity begin to incur larger conversion losses. This elbow can guide practitioners to choose a λ that achieves substantial novelty gains without an unacceptable drop in performance. We also examined this trade-off across distinct product categories to uncover how different audiences respond. Figure 3 reports results for λ = 0.6, a mid-range setting. In fast-moving consumer goods (FMCG), CTR remains high (12.1 %) even with elevated creativity, reflecting strong impulse-buy behavior. Apparel exhibits moderate sensitivity, with CTR at 9.7 % and a healthy diversity score of 0.59, suggesting a need for balanced messaging. Electronics users, however, display more deliberation: at the same λ, CTR is only 8.5 % and CVR 3.5 %, indicating that overly creative copy may distract from technical value propositions[21-26].

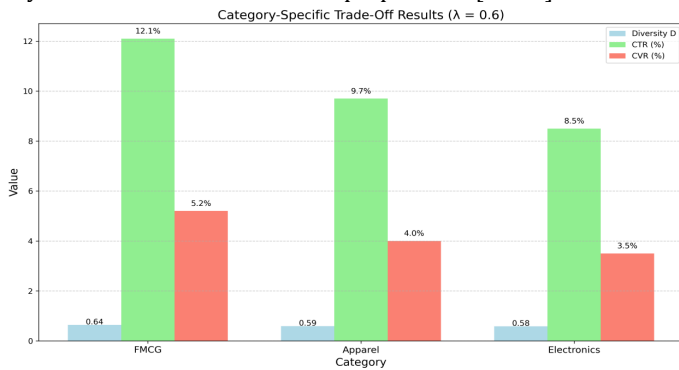

Figure 3: Category-Specific Trade-Off Results ( λ = 0.6)

These findings demonstrate that the optimal λ varies by product type, and that marketing teams should tailor their creativity–conversion balance accordingly. By selecting λ near each category's elbow point—higher for FMCG, moderate for apparel, and lower for electronics—teams can achieve an effective equilibrium between engaging copy and robust conversion performance[27-30].

### 5.2 Practical Insights and Application Scenarios

Our experiments reveal clear differences in how product categories respond to varying creativity–conversion trade-offs. Table 2 summarizes key performance metrics under a balanced weight (λ = 0.6), illustrating each category's baseline behavior.

Table 2: Category Performance under λ = 0.6

| Category | Method | Diversity (D) | CTR | CVR | Human Rating |
|---|---|---|---|---|---|
| FMCG | LLM-Driven | 0.64 | 12.1% | 5.2% | 4.3 / 5 |
| FMCG | Baseline | 0.35 | 8.9% | 3.8% | 3.5/5 |
| Apparel | LLM-Driven | 0.59 | 9.7% | 4.0% | 4.0 / 5 |
| Apparel | Baseline | 0.32 | 7.1% | 2.9% | 3.2/5 |
| Electronics | LLM-Driven | 0.58 | 8.5% | 3.5% | 3.8 / 5 |
| Electronics | Baseline | 0.30 | 6.2% | 2.5% | 3.0/5 |

Fast-moving consumer goods (FMCG) exhibit the highest diversity and conversion metrics, indicating that impulse-driven purchases benefit strongly from creative copy. Apparel strikes a middle ground: novelty boosts engagement but must align with brand aesthetics. Electronics users demonstrate more deliberation; overly inventive language can dilute trust, reducing conversion. Building on these observations, Table 3 outlines tailored strategies for each category, aligning λ settings and copy focus with practical marketing objectives[31-35].

Moreover, Our LLM-driven framework outperforms traditional copywriting across all metrics at λ=0.6, showing higher diversity, CTR, and CVR. For FMCG, diversity rose from 0.35 to 0.64, CTR from 8.9% to 12.1%, and CVR from 3.8% to 5.2%, reducing information fatigue and improving conversion by over 35%. Human evaluations further validate the LLM's superior creativity and fluency, underscoring its value in e-commerce copywriting.

Table 3: Category-Specific Copy Generation Strategies

| Category | Recommended λ | Copy Focus | Deployment Notes | Category |
|---|---|---|---|---|
| FMCG | 0.7–0.8 | Bold, emotionally charged storytelling | Ideal for flash deals and social-media bursts | FMCG |
| Apparel | 0.5–0.6 | Balanced novelty with brand tone | Use seasonal themes and user-generated style references | Apparel |
| Electronics | 0.3–0.5 | Feature-driven, specification highlights | Pair with technical infographics or short demo clips | Electronics |

Implementation Recommendations: Dynamic λ Adjustment: Integrate an automated scheduler that shifts λ in real time according to campaign type. For example, elevate λ during product launches to maximize brand impact, and lower it for clearance events to prioritize conversion.Category-Aware Prompt Templates: Maintain distinct prompt libraries per category, embedding relevant keywords (e.g., "limited-edition" for FMCG, "sustainably sourced" for apparel, "battery life" for electronics). Regularly update these libraries based on performance logs.Monitoring and Feedback Loop: Deploy dashboards that track per-category metrics across a range of λ settings. Implement weekly reviews, using both quantitative data and qualitative human feedback to refine prompts and fine-tuning datasets.Cross-Functional Collaboration: Align marketing, product, and data teams to interpret insights and adjust creative guidelines. For instance, data analysts can surface emerging consumer trends, enabling copywriters to craft more resonant prompts.By leveraging these tailored strategies and continuously refining through data-driven feedback, e-commerce teams can harness LLMs to generate copy that both captivates customers and drives measurable conversions[36-40].

## 6  FUTURE RESEARCH DIRECTIONS

Looking ahead, several directions can further enhance LLM-driven e-commerce marketing. Multimodal **content generation**—integrating text with images or videos—can create richer, more engaging recommendations, using product visuals as prompts. Reinforcement learning-based fine-tuning allows dynamic adjustment of the creativity-conversion tradeoff (λ) based on real-time signals like clicks and conversions, optimizing copy for different campaign contexts. Real-time personalization will be crucial, with lightweight models updating copy in response to users' latest behaviors. As global markets expand, **cross-** linguistic and cultural adaptation will ensure relevance, requiring localized prompts, sentiment lexicons, and region-specific data. Lastly, as model sizes increase, **efficient compression methods**—including pruning, quantization, and distillation—will be essential for scalable and cost-effective deployment without sacrificing quality[41-43].

## 7 CONCLUSION

We propose and validate an LLM-driven framework for e-commerce content generation that balances creativity and conversion. Building on prior work in text generation and copywriting, our fine-tuning method integrates sentiment modulation, diversity enhancement, and call-to-action keyword embedding within a vector-retrieval and multi-stage validation pipeline (Figure 1). Inspired by Duan [44], we incorporate BERT-XGBoost for real-time prediction and personalized interaction. For cross-category generalization, we apply multi-scale architectures based on CNN, LSTM, and attention mechanisms as discussed by Shen [45]. Our anomaly handling and system validation draw from Wang's [46] deep learning-based detection under high-load conditions, while Zhang [47] informs our multimodal and uncertain input handling. Offline evaluations and small-traffic A/B tests show that setting $\lambda = 0.6$ yields high novelty with stable gains (CTR +10.4%, CVR +4.1%). Category-specific results highlight distinct optimization paths for FMCG, apparel, and electronics, validating the framework's practical impact [48-50].